\documentclass[letterpaper]{article} 
\usepackage{aaai2026}  
\usepackage{times}  
\usepackage{helvet}  
\usepackage{courier}  
\usepackage[hyphens]{url}  
\usepackage{graphicx} 
\usepackage{natbib}  
\usepackage{caption} 
\usepackage{algorithm}
\usepackage{algorithmic}

\usepackage{newfloat}
\usepackage{listings}
\DeclareCaptionStyle{ruled}{labelfont=normalfont,labelsep=colon,strut=off} 
\floatstyle{ruled}
\newfloat{listing}{tb}{lst}{}
\floatname{listing}{Listing}
\usepackage{times}
\usepackage{soul}
\usepackage{url}
\usepackage[utf8]{inputenc}
\usepackage{amsmath}
\usepackage{amsthm}
\usepackage{booktabs}
\usepackage{algorithm}
\usepackage{algorithmic}
\usepackage[switch]{lineno}

\newcommand{\algo}{MrCoM} 

\usepackage{graphicx}
\usepackage{latexsym}
\usepackage{amsmath}
\usepackage{algorithm}
\usepackage{algorithmic}
\usepackage{xcolor}
\usepackage{amssymb}
\usepackage{multirow}
\usepackage{siunitx}
\usepackage{booktabs}
\usepackage{subcaption}

\newtheorem{theorem}{Theorem}
\newtheorem{lemma}[theorem]{Lemma}

\newtheorem{assumption}[theorem]{Assumption}

\title{MrCoM: A Meta-Regularized World-Model Generalizing Across Multi-Scenarios}

\author{
Xuantang Xiong\textsuperscript{\rm 1,4,5}\equalcontrib, Ni Mu\textsuperscript{\rm 2}\equalcontrib, Runpeng Xie\textsuperscript{\rm 1}, Senhao Yang\textsuperscript{\rm 1}, Yaqing Wang\textsuperscript{\rm 1}, Lexiang Wang\textsuperscript{\rm 1}, \\Yao Luan\textsuperscript{\rm 2}, Siyuan Li\textsuperscript{\rm 3}, Shuang Xu\textsuperscript{\rm 1}, Yiqin Yang\textsuperscript{\rm 1}\thanks{Corresponding Authors.}, Bo Xu\textsuperscript{\rm 1}\footnotemark[2]\\
}
\affiliations{
    \textsuperscript{\rm 1}The Key Laboratory of Cognition and Decision Intelligence for Complex Systems,\\ Institute of Automation, Chinese Academy of Sciences, Beijing, China\\
    \textsuperscript{\rm 2}Department of Automation, Tsinghua University,
    \textsuperscript{\rm 3}Faculty of Computing, Harbin Institute of Technology\\
    \textsuperscript{\rm 4}Tencent Robotics X \& Futian Laboratory, Shenzhen\\
    \textsuperscript{\rm 5}Shenzhen Institutes of Advanced Technology, Chinese Academy of Sciences
}

\usepackage{bibentry}

\begin{document}

\maketitle

\begin{abstract}
Model-based reinforcement learning (MBRL) is a crucial approach to enhance the generalization capabilities and improve the sample efficiency of RL algorithms. 
However, current MBRL methods focus primarily on building world models for single tasks and rarely address generalization across different scenarios.
Building on the insight that dynamics within the same simulation engine share inherent properties, we attempt to construct a unified world model capable of generalizing across different scenarios, named Meta-Regularized Contextual World-Model (MrCoM).
This method first decomposes the latent state space into various components based on the dynamic characteristics, thereby enhancing the accuracy of world-model prediction.
Further, MrCoM adopts meta-state regularization to extract unified representation of scenario-relevant information, and meta-value regularization to align world-model optimization with policy learning across diverse scenario objectives.
We theoretically analyze the generalization error upper bound of MrCoM in multi-scenario settings.
We systematically evaluate our algorithm's generalization ability across diverse scenarios, demonstrating significantly better performance than previous state-of-the-art methods.
\end{abstract}

\section{Introduction}
Reinforcement learning, a key algorithm for solving decision-making problems, has achieved significant success in domains such as games~\cite{silver2018general,berner2019dota}, robotic control~\cite{akkaya2019solving}, autonomous driving~\cite{kiran2021deep}, and even large language model tuning~\cite{sun2023aligning}. 
One crucial requirement for broader RL applications is to improve the sample efficiency, and recent advances in model-based approaches have emerged as a promising solution by learning a world-model.

Most model-based RL methods mainly focus on single-task settings, where a dedicated world model is trained for each individual task. 
Recent studies have demonstrated the robustness of world model hyperparameters across diverse scenarios, that is, maintaining identical training configurations to train world models under varying conditions.
However, they seldom investigate cross-scenario generalization of the world models themselves. This critical gap prevents the deployment of a unified world model across multiple environments, which could substantially reduce training costs.
Therefore, this naturally leads to the following question:
\begin{center}
    \it{How to learn a unified world-model capable of generalizing across multi-scenarios?}
\end{center}

In this work, we aim to provide an effective solution to construct the unified world-model. 
First, we decompose the latent state space into various components based on the dynamic characteristics, thereby enhancing the accuracy of world-model prediction.
Further, we propose the meta-state regularization mechanism to extract unified representation of scenario-relevant information, and meta-value regularization to align world-model optimization with policy learning across diverse scenario objectives. 
We name our method as Meta-Regularized Contextual World-Model (\algo).
We conduct the theoretical analysis for the model's generalization capability and derive the upper bound of the generalization error of our method.


For the experiments, we select the Mujoco-based MDC~\cite{tassa2018deepmind,todorov2012mujoco}, a simulation engine for robotic actions, as the benchmark.
We construct the meta-scenarios set by changing the scenario dynamics and objective (e.g., robot's limb size and length).
Through extensive experimental results, we demonstrate the advantage of our method over current state-of-the-art approaches. 

In summary, our contributions are threefold:
%
\begin{enumerate}
    \item We propose \algo~to learn a unified world-model based on the meta-state and meta-value regularization mechanism.
    \item We conduct the theoretical analysis for \algo, deriving the generalization error upper bound.
    \item We conduct comprehensive experiments demonstrating that, compared to other baselines, the world model learned by \algo~can generalize across diverse scenarios.
\end{enumerate}

\begin{figure*}[t]
\centering
\includegraphics[width=1.0\textwidth]{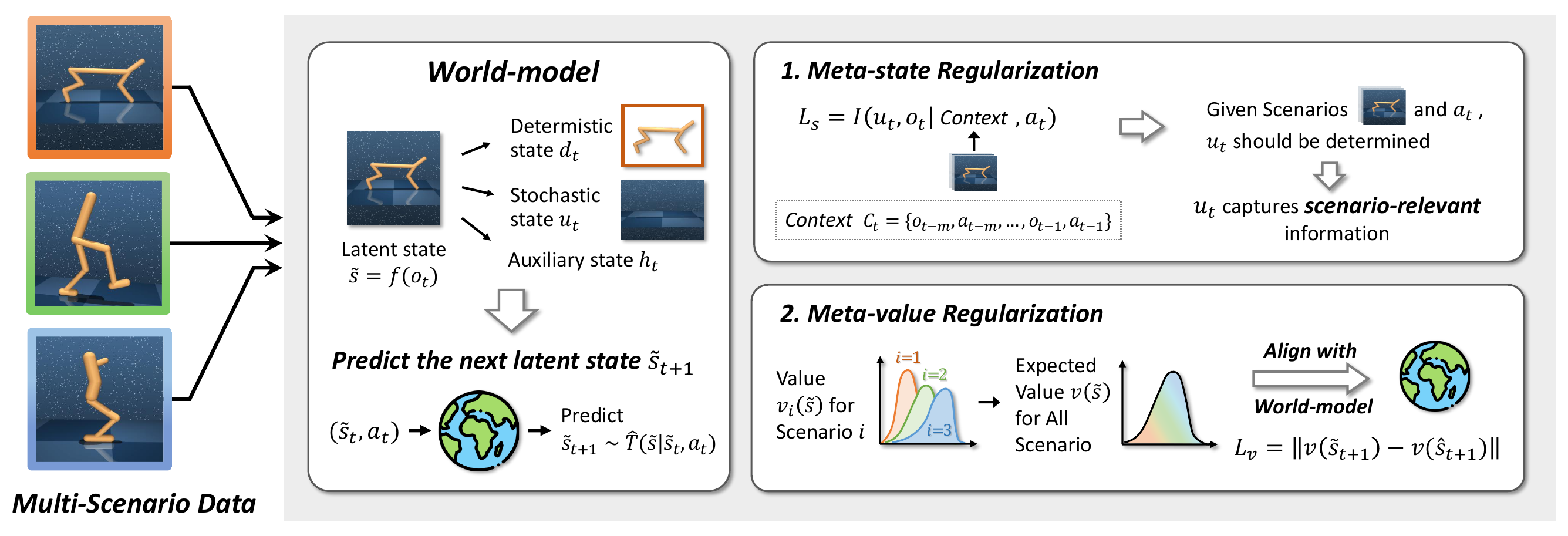}
\caption{Framework of \algo, which merges multi-scenario data into a unified world-model. 
The meta-state regularization extracts
scenario-relevant information, and meta-value regularization aligns world-model optimization with policy learning.}
\label{fig:framework_illustration}
\end{figure*}

\section{Problem Formulation}

\subsection{POMDP} 
A Partially Observable Markov Decision Process (POMDP) is an extension of a Markov Decision Process (MDP) used to model decision-making problems under uncertainty and incomplete information. It can be represented as a tuple $\mathcal{M}=\{\mathcal{S}, \mathcal{A}, r, T, {\rho}_0, \gamma, \mathcal{O}, \Omega \}$.
Here, $\mathcal{S} \in \mathbb{R}^n$ denotes the state space, $\mathcal{A} \in \mathbb{R}^m$ represents the action space, $r: \mathcal{S} \times \mathcal{A} \times \mathcal{S} \rightarrow \mathbb{R}$ is the reward function, $T: \mathcal{S} \times \mathcal{A} \rightarrow \mathcal{S}$ defines the dynamics of state transitions, ${\rho}_0 \in \mathcal{S}$ represents the initial state distribution, $\gamma \in [0,1)$ is the discount factor, $\mathcal{O} \in \mathbb{R}^k$ denotes the observation space, and $\Omega: S \rightarrow \mathcal{O}$ is observation function representing the mapping from states to observations.

The policy of an agent, denoted as $\pi_\phi: \mathcal{O} \rightarrow \mathcal{A}$, is parameterized by $\phi$. 
The discounted infinite horizon return is defined as $G_{\pi} = \sum^\infty_{t=0} \gamma^{t} r(s_t, a_t)$, where $t$ represents the time step and $a$ denotes the action taken by the agent. 
The objective is to find the optimal policy parameter $\phi^{*}$ that maximizes the cumulative discounted reward:
\begin{align}\label{qu1}
    \phi^{*} = \arg \max_{\phi} \mathbb{E} \left[ \sum^\infty_{t=0} \gamma^{t} r(s_t, \pi_\phi(o_t)) \right]
\end{align}

\subsection{Meta-POMDP}
In this work, we consider finding a policy that maximizes expected return in a Meta-POMDP, defined as $\mathcal{M}= \{\mathcal{M}^i\}_{i=1}^N = \{\mathcal{S}^i, \mathcal{A}^i, r^i, T^i, {\rho}_0^i, \gamma, \mathcal{O}^i, \Omega^i \}_{i=1}^N$, with a set of scenario indices $1, \cdots, N$.
Each scenario $i$ is randomly sampled from the meta-scenario set $\mathcal{T}^i \sim p(\mathcal{T})$, where $\mathcal{T}^i$ corresponds to a particular POMDP tuple $\mathcal{M}^i$.
In this work, we assume that each scenario $i$ presents a different state space $\mathcal{S}^i$, action space $\mathcal{A}^i$, reward function $r^i$, dynamics $T^i$ and observation space $\mathcal{O}^i$.
We formulate the policy optimization problem as finding a policy that maximizes expected return over all the scenarios:

\begin{align}\label{qu1}
    \phi^{*} = \arg \max_{\phi} \mathbb{E}_{\mathcal{T}^i \sim p(\mathcal{T})}\left[ \sum^\infty_{t=0} \gamma^{t} r^i(s^i_t, \pi_{\phi}(o^i_t)) \right]
\end{align}

\paragraph{Comparison with other setting:}
Unlike prior works, we aim to learn a unified world-model effective across varying state spaces $\mathcal{S}^i$, action spaces $\mathcal{A}^i$, dynamics $T^i$, reward functions $r^i$, and observation functions $\mathcal{O}^i$.
For example, CaDM~\cite{lee2020context} seeks to construct a world-model capable of generalizing across diverse dynamics $T^i$, which is represented as $\{\mathcal{M}^i\}_{i=1}^N = \{\mathcal{S}, \mathcal{A}, r, T^i, {\rho}_0, \gamma, \mathcal{O}, \Omega \}_{i=1}^N$.
Dreamerv3~\cite{hafner2023mastering} is designed to construct a world-model for individual tasks, where $\mathcal{M} = \{\mathcal{S}, \mathcal{A}, r, T, {\rho}_0, \gamma, \mathcal{O}, \Omega \}$.
MAMBA~\cite{rimon2024mamba} seeks to construct a world model effective across diverse scenarios with varying dynamics $T^i$ and reward functions $r^i$, which is denoted as $ \{\mathcal{M}^i\}_{i=1}^N = \{\mathcal{S}, \mathcal{A}, r^i, T^i, {\rho}_0, \gamma, \mathcal{O}, \Omega \}_{i=1}^N$.

\subsection{Model-based RL}
Model-Based Reinforcement Learning (MBRL) constructs an approximate dynamics model $\hat{T}_\theta (s^\prime_t\mid s_t,a_t)$, where the model parameters $\theta$ are optimized based on observed data $\mathcal{D}$ through maximizing the likelihood:
\begin{align}\label{model_learning}
    \theta^* = \arg \max_{\theta} \sum_{(s, a, s^\prime) \sim \mathcal{D}} \log \hat{T}_\theta(s^\prime_t\mid s_t, a_t)
\end{align}

In the MBRL framework, policy learning can take advantage of real data and simulated data generated by the approximated dynamics model $\hat{T}_\theta (s^\prime_t\mid s_t,a_t)$, which enhances sample efficiency and reduces dependency on real environment interactions.


\begin{figure*}[t]
\centering
\begin{subfigure}[b]{0.85\linewidth}
    \includegraphics[width=\columnwidth]{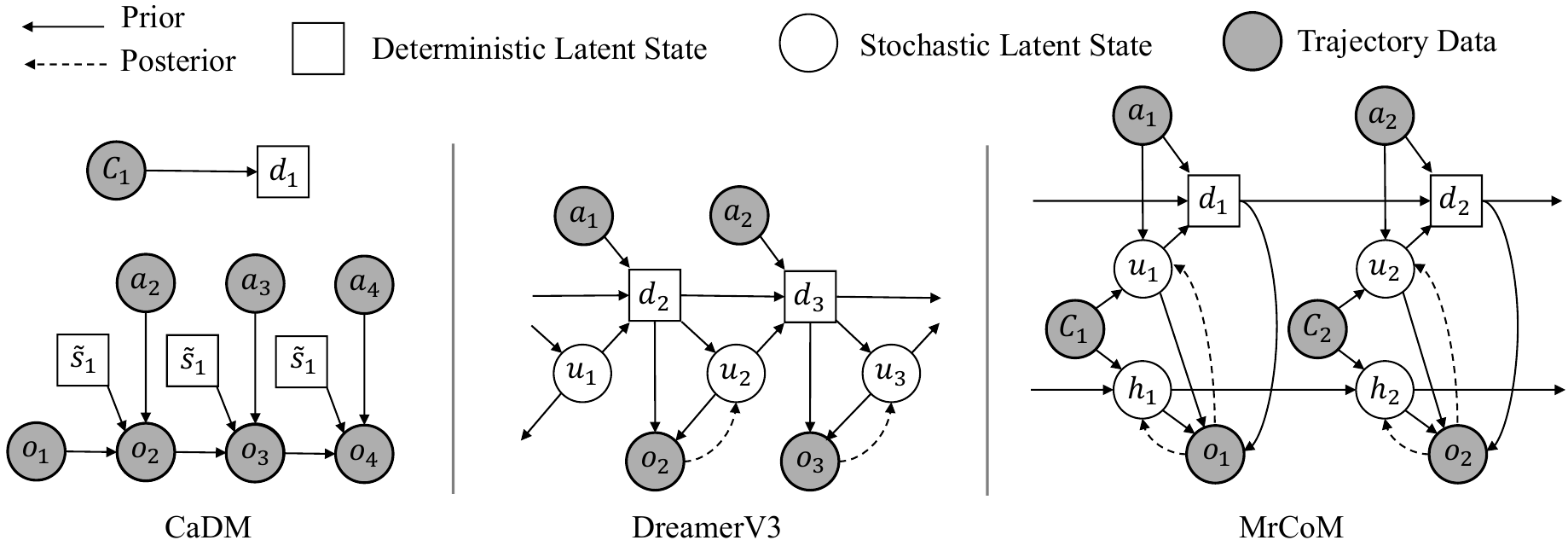}
\end{subfigure}
\caption{Model architecture of algorithms. Compared to other methods, \algo~implements a refined partitioning of the latent state space. In this Figure, $\tilde{s}$ denotes the latent state, $d$ the deterministic latent state, $u$ the stochastic latent state, and $h$ the auxiliary state.
In \algo, the latent state $\tilde{s}_t$ is composed of the concatenation of $u_t$, $d_t$, and $h_t$.}
\label{fig:s-mrcom}
\end{figure*}

\section{Method}
In this section, we propose a novel method to learn a unified world-model capable of generalizing across multi-scenarios, named \textbf{M}eta-\textbf{R}egularized \textbf{C}ontextual W\textbf{o}rld-\textbf{M}odel~(MrCoM).
We first present the model architecture, which decomposes the latent state space based on dynamic characteristics, enhancing prediction accuracy of the world model. 
Subsequently, we separately introduce meta-state regularization and meta-value regularization mechanisms.
These two mechanisms extract scenario-relevant information and align model-policy optimization, achieving cross-scenario generalization of the world model.
The framework of \algo~is shown in Figure~\ref{fig:framework_illustration}.


To optimize the world-model across multiple scenario, we inject contextual information into state prediction with the transformer architecture. 
During inference initialization, the contextual information will activate the model's domain-specific predictive capability.
The contextual information is defined as:
\begin{align}
C_t = \{o_{t-m}, a_{t-m}, \ldots, o_{t-1}, a_{t-1}\},
\end{align}
where $m$ denotes the context length. 
When $t < m$, scenario-related expert trajectories are used as contextual information, with the first-in-first-out mechanism to update the generated latest trajectories.

\subsection{Model Architecture}
\label{sec: model architec}
Complex environments often exhibit stochastic disturbances, making precise modeling challenging. 
To tackle this issue, we first assume a mapping function $f^\star$ that projects environmental states into a unified latent space $\widetilde{\mathcal{S}}$:

\begin{assumption}[Unified Mapping]
\label{assump1}
For all scenarios, states and actions can be mapped by a function $f^\star$ to a unified latent state space $\widetilde{\mathcal{S}}$. For simplicity, the unified mapping of the action space is not explicitly expressed here. Formally, the state unified mapping can be described as:
\begin{equation}
\begin{aligned}\label{eq1}
& \forall \mathcal{T}^i \sim p(\mathcal{T}), \quad \exists f^\star: \mathcal{O}^i \to \widetilde{\mathcal{S}} \\ 
\text{s.t., } & f^\star(o) = \tilde{s}, \quad \forall o \in \mathcal{O}^i, \text{ where } \tilde{s} \in \widetilde{\mathcal{S}}.
\end{aligned}
\end{equation}
\end{assumption}

We assume the unified latent space $\widetilde{\mathcal{S}}$ can be decomposed into stochastic $u_t$ and deterministic $d_t$ components.
We derive the stochastic component $u_t$ from contextual information $C_t$, which follows the Gaussian distribution to capture the uncertainty in state transitions:
\begin{equation}
\begin{aligned}
p^{u_t}_\theta &= p_\theta(u_t\mid C_t, a_t)
\end{aligned}
\end{equation}
We compute the deterministic component $d_t$ based on historical information $d_{t-1}$ and the current stochastic input $u_t$,  thereby preserving historical stochastic information:
\begin{equation}
\begin{aligned}
p_\theta^{d_t} &= p_\theta(d_t\mid d_{t-1}, u_{t}, a_t)
\end{aligned}
\end{equation}
Since we aim to learn a unified world model applicable across multiple scenarios, we introduce an additional auxiliary state $h_t$ conditioned on contextual information $C_t$ to enhance prediction accuracy and minimize information loss.
This auxiliary state $h_t$ follows a Gaussian distribution:
\begin{equation}
    \begin{aligned}
        p^{h_t}_\theta &= p_\theta(h_t\mid C_{t}, h_{t-1})
    \end{aligned}
\end{equation}

Based on the above analysis, the latent state $\tilde{s}_t$ is composed of the concatenation of $u_t$, $d_t$, and $h_t$.
For the posterior distribution, we utilize the original observation $o_t$ as the additional input and reconstruct the original observation $o_t$ conditioned on the stochastic component $u_t$, deterministic component $d_t$, and auxiliary state $h_t$:
\begin{equation}
\begin{aligned}
q^{u_t}_\theta &= q_\theta(u_t\mid C_{t}, a_{t}, o_t) \\
q^{h_t}_\theta &= q_\theta(h_t\mid C_{t}, h_{t-1}, o_t)\\
p^{o_t}_\theta &= p_\theta (o_t\mid u_t, d_t, h_t)
\end{aligned}
\end{equation}
Based on the above analysis, our world-model $\hat{T}_{\theta}$ comprises the prior encoders $p^{u_t}_\theta, p^{d_t}_\theta, p^{h_t}_\theta$, the posterior encoders $q^{u_t}_\theta, q^{h_t}_\theta$, and the reconstruction decoder $p^{o_t}_\theta$.
Based on the variational bound, we have the following loss function:
\begin{equation}
\begin{aligned}
\mathcal{L}_\text{var} &=  \mathcal{L}_\text{KL} + \mathcal{L}_\text{recon} \\
&=  \mathbb{E}_{\mathcal{D}} \left[ \text{KL}(p_\theta^{u_t} \|q_\theta^{u_t}) +\text{KL}(p_\theta^{h_t}\| q_\theta^{h_t}) -\log p^{o_t}_\theta)\right],
\label{eq: var loss}
\end{aligned}
\end{equation}
where $\mathcal{L}_\text{KL}$ denotes the KL divergence between the prior and posterior distributions of latent states, and $\mathcal{L}_\text{recon}$ represents the reconstruction error of original observations.
Please refer to Figure~\ref{fig:s-mrcom} for the model architecture.

\subsection{Meta-State Regularization}
\label{sec: meta state}
Original observations $o_t$ typically contain both scenario-relevant and scenario-irrelevant information. Scenario-relevant information refers to state variables determined by actions, whereas scenario-irrelevant information denotes action-invariant state elements. Crucially, scenario-relevant information correlates with scenario objectives and plays a vital role when the world model operates across various domains.
Therefore, to learn a unified world-model applicable across multiple scenarios, we must extract scenario-relevant information from original observations.

Inspired by Denoised MDPs, we find that the transition of scenario-relevant states can be largely determined given current actions $a_t$ and contextual information $C_t$. Consequently, to filter out scenario-irrelevant information from original observations, we optimize the world-model by minimizing mutual information $I(u_t, o_t \mid C_t, a_t )$ between stochastic components $u_t$ and original observations $o_t$.
Specifically, we can optimize the upper bound for the mutual information by introducing an additional variational approximation distribution $q(u_t|C_t, a_t)$~\cite{poole2019variational}:
\begin{equation}
\begin{aligned} 
    I&(u_t , o_t  \mid C_t, a_t )  \equiv \mathbb{E}_{p(C_t, a_t)} \mathbb{E}_{p(u_t, o_t)}\left[\log \frac{p(u_t | o_t, C_t, a_t)}{p(u_t|C_t, a_t)}\right] \\
    & =\mathbb{E}_{p(C_t, a_t)} \mathbb{E}_{p(u_t, o_t)}\left[\log \frac{p(u_t | o_t, C_t, a_t) q(u_t|C_t, a_t)}{q(u_t|C_t, a_t) p(u_t|C_t, a_t)}\right] \\
    & =\mathbb{E}_{p(C_t, a_t)} \mathbb{E}_{p(u_t, o_t)}\left[\log \frac{p(u_t |C_t, a_t, o_t)}{q(u_t|C_t, a_t)}\right] \\
    & \qquad \qquad \qquad \qquad - \text{KL} \left(p(u_t|C_t, a_t) \| q(u_t|C_t, a_t)\right) \\
    & \leq \mathbb{E}_{p(C_t, a_t)} \mathbb{E}_{p(o_t)}\left[ \text{KL} \left(p(u_t | C_t, a_t, o_t, ) \| q(u_t | C_t, a_t)\right)\right]
\end{aligned}
\end{equation}
When the prior distribution approaches the posterior distribution, we have $\text{KL} \left(p_{\theta}(u_t \mid C_t, a_t, o_t) \| q_{\theta}(u_t \mid C_t, a_t)\right) \approx \text{KL} \left(p_{\theta}(u_t \mid C_t, a_t) \| q_{\theta}(u_t \mid C_t, a_t, o_t)\right)$, and this divergence serves as a tight upper bound on mutual information. 
Therefore, based on the above analysis, we can minimize the following loss to extract scenario-relevant information into world-model: 
\begin{equation}
    \mathcal{L}_s = \mathbb{E}_{\mathcal{D}} \left[\mathrm{KL} \left(p_{\theta}(u_t \mid C_t, a_t) \| q_{\theta}(u_t \mid C_t, a_t, o_t)\right)\right]
    \label{ads}
\end{equation}



\subsection{Meta-Value Regularization}
\label{sec: meta value}
In the multi-scenario setting, the vast state space makes it challenging for the world-model to achieve accurate predictions universally.
To address this issue, we let the world model focus primarily on regions involved in value function updates, which enhances the precision of value estimation and thereby improves policy performance.

To this end, we propose the meta-value regularization mechanism to align world-model optimization with policy learning.
Specifically, we define the meta-value $v_\psi(\tilde{s}_t)$, which refers to the shared value extracted from latent states $\tilde{s}_t$ across various scenario distributions. 
We optimize $v_\psi(\tilde{s}_t)$ via the following loss function:
\begin{equation}
\label{equ:loss2}
\mathcal{L}_{\text{value}}  = \mathbb{E}_{\mathcal{T}^i \sim p(\mathcal{T})} \left[\left\| v_{\psi_i}(\tilde{s}_t) - v_\psi(\tilde{s}_t) \right\|^2\right],
\end{equation}
where $v_{\psi_i}(\tilde{s}_t)$ represents the value function for a specific scenario $\mathcal{T}^i$ and $\tilde{s}_t=\{u_t,d_t,h_t\}$.
We update $v_{\psi_i}$ based on the following Bellman equation:
\begin{equation}
\label{equ:loss1}
\begin{aligned}
\mathcal{L}_{\text{value}_i}  =\mathbb{E}_{\mathcal{T}^i \sim p(\mathcal{T})}  \left[ \left\| v_{\psi_i}(\tilde{s}_t) - \left( r_t + \gamma v_{\psi_i}(\tilde{s}_{t+1}) \right) \right\|^2\right].
\end{aligned}
\end{equation}

Further, we update the world-model based on the learned meta-value $v_\psi(\tilde{s}_t)$:
\begin{equation}
    \mathcal{L}_{v} = \mathbb{E}_{\mathcal{T}^i \sim p(\mathcal{T})} \left[\left\| v_\psi(\tilde{s}_{t+1}) - \hat{T}_\theta (\hat{s}_{t+1}|\tilde{s}_t,a_t) v_\psi(\hat{s}_{t+1}) \right\|\right],
\end{equation}
where $\hat{s}_{t+1}$ represents the latent state predicted by the world-model $\hat{T}_\theta$. 

\subsection{Implementation Details}
Based on the analysis in Section~\ref{sec: model architec},~\ref{sec: meta state},~\ref{sec: meta value}, the total loss function of~\algo~is defined as follows:
\begin{equation}\label{loss:mrcom}
    \mathcal{L}_\text{MrCoM} = \lambda_\text{var}\mathcal{L}_\text{var}+\lambda_s\mathcal{L}_{s}+\lambda_v\mathcal{L}_{v}.
\end{equation}
where $\lambda_\text{var}$, $\lambda_s$, $\lambda_v$ are hyperparameters that balance the weights of the loss terms.

The overall procedure of the algorithm comprises two phases: world-model training and scenario adaptation. 
In the training phase, we collect data across diverse scenarios using the behavioral policy $\pi_{\phi_i}$, followed by training the world model according to Equation~\ref{loss:mrcom}. 
During scenario adaptation, given target scenario $\mathcal{T}^{\text{target}}$, we perform rollouts based on the trained world model $\hat{T}_{\theta}$ to augment the dataset, which is then used to train a standard RL algorithm.
The detailed algorithmic procedure and hyperparameters are provided in Appendix~\ref{app: alg} and Appendix~\ref{appendix: details}.

\section{Theoretical Analysis}
This section establishes the theoretical foundation of \algo's framework and provides insights into the proposed objective. 
Firstly, we introduce the following assumption:

\begin{assumption}[Dynamics Homogeneity]
\label{assump2}
For states from different scenarios that map to the same state $\tilde{s}_t \in \widetilde{\mathcal{S}}$, their state transition probabilities under the same action are identical. This is represented as:
\begin{equation}
\begin{aligned}\label{eq2}
& \text{if}\quad f^\star(o^i_t) = f^\star(o^j_t) = \tilde{s}_t, \quad \forall o^i_t \in \mathcal{O}^i, \forall o^j_t \in \mathcal{O}^j,\\
&\text{then} \quad T^i(s^i_t, a_t) = T^j(s^j_t, a_t), \quad \forall a_t \in \mathcal{A}.
\end{aligned}
\end{equation}
\end{assumption}
Under the Assumption~\ref{assump1} and \ref{assump2}, it can be inferred that there exists a shared latent transition function $\widetilde{T}$ across scenarios, defined on the latent space $\widetilde{S}$:
\begin{equation}
\begin{aligned}
& \widetilde{T}(\tilde{s}'\mid \tilde{s}, a) = T^i(s'\mid s, a) \\
& \forall \mathcal{T}^i \sim p(\mathcal{T}), s \in \mathcal{S}^i, a \in \mathcal{A}^i, \exists \widetilde{T}: \widetilde{\mathcal{S}} \times \mathcal{A} \to \widetilde{\mathcal{S}},
\end{aligned}
\end{equation}
where $T^i$ is the true transition function on the scenario $i$.
Then, we define the dynamic error as follows:
\begin{equation}
    \max _{i} \mathbb{E}_{\mathcal{T}^i \sim p(\mathcal{T})} D_{\text{TV}}\left(\widetilde{T} \left(\tilde{s}^{\prime} \mid \tilde{s},  a \right) \|  T^{i}\left(s^{\prime} | s, a\right)  \right) \leq \epsilon_{T}
\end{equation}

Next, we define the state representation error as follows:
\begin{equation}
    \max_{i} \mathbb{E}_{\mathcal{T}^i \sim p(\mathcal{T})}  \left| f(o) - \tilde{s}  \right|    \leq \epsilon_{S},
\end{equation}
where $f$ is the representation function and $\tilde{s}$ is the ture latent state.
Based on the above definitions, we first consider the dynamics model error bound under the state representation:
\begin{lemma}~\label{state_dyna}
Given the state representation error $\epsilon_{S}$ and dynamics model error $\epsilon_{T}$, the upper bound of the dynamics error under the state representation is:
\begin{equation}
D_{\text{TV}}\left( \widetilde{T} \left(f(o^\prime) \mid f(o), a \right) \| T^{i}\left(s^{\prime} \mid s, a\right)\right) \leq \epsilon_T + C_{T} \cdot \epsilon_{S}.
\end{equation}
Here, $C_T = \max _s \nabla_s \sum_a T(s^\prime \mid s, a)$ denotes the maximum derivative of the dynamics function concerning $s$, representing the sensitivity of state changes to the state transition function.
\end{lemma}
\begin{proof}
    Please refer to Appendix~\ref{proof: lemma3} for the detailed proof.
\end{proof}

Then, we derive the policy difference bound under state representation as follows:
\begin{lemma}~\label{state_pi}
Given policy difference in the state space $\max_{s} D_{TV}\left(\pi_{1}(a \mid s) \| \pi_{2}(a \mid s)\right) \leq \epsilon_{\pi}$, the upper bound of the policy difference under the state representation error $\epsilon_{S}$ is:
\begin{equation}
    D_{T V}\left(\pi_{1}(a \mid f(o)) \| \pi_{2}(a \mid s)\right) \leq \epsilon_{\pi}+\frac{1}{2} C_{\pi} \cdot \epsilon_{S}
\end{equation}
Here, $C_{\pi} = \max _s \nabla_s \sum _a \pi_1(a|s)$ represents the maximum sensitivity of the policy function to state variations. 
\end{lemma}
\begin{proof}
    Please refer to Appendix~\ref{proof: lemma4} for the detailed proof.
\end{proof}

Next, we consider the upper bound of the performance difference under two distinct dynamics and policies as follows:
\begin{lemma}\label{dyna_pi}
Given dynamics model error $\epsilon_T$, the policy difference $\epsilon_\pi$ and the maximum reward $R$, the performance difference has the following upper bound:
\begin{equation}
    \left|G^1(\pi_1)-G^2(\pi_2)\right| \leq \frac{2 R \gamma\left(\epsilon_{\pi}+\epsilon_{T}\right)}{(1-\gamma)^{2}}+\frac{2 R \epsilon_{\pi}}{1-\gamma}
\end{equation}
Here, $G^1(\pi_1)$ denotes the performance of $\pi_1$ under dynamics $T^1$, and $G^2(\pi_2)$ denotes the performance of $\pi_2$ under dynamics $T^2$. 
\end{lemma}
\begin{proof}
    Please refer to Appendix~\ref{proof: lemma5} for the detailed proof.
\end{proof}

Let $\widetilde{G}^i(\pi)$ denote the performance of $\pi(f(o))$ under the dynamics $\widetilde{T}^i(f(o), a)$.
Let $\widetilde{G}_\theta(\pi)$ denote the performance under the learned world-model $\widetilde{T}_{\theta}$.
By combining Lemma~\ref{state_dyna} and Lemma~\ref{state_pi} with Lemma~\ref{dyna_pi}, 
we can derive the generalization error bound in the Meta-POMDP setting:

\begin{theorem}\label{theorem2}
Given dynamics model error $\epsilon_T$, policy difference $\epsilon_\pi$, and the state representation error $\epsilon_{S}$, the upper bound of the generalization error on the Meta-POMDP is:
\begin{align}
    \left|\widetilde{G}^i(\pi)-\widetilde{G}_\theta(\pi)\right| \leq  \frac{ R\gamma[4\epsilon_{\pi}+ 2\epsilon_{T} +(C_\pi +2C_T) \epsilon_{S} ] }{(1-\gamma)^{2}}  \notag \\
        +\frac{2R (2\epsilon_{\pi}+ C_\pi \epsilon_{S})}{1-\gamma}.
\end{align}
\end{theorem}

\begin{proof}
    Please refer to Appendix~\ref{proof theorem 2} for the detailed proof.
\end{proof}

From the above theory, the upper bound comprises three primary error sources: dynamics model error $\epsilon_T$, state representation error $\epsilon_{S}$, and policy difference $\epsilon_\pi$. 
We find that these three errors contribute linearly to the generalization error bound of the model. 
In \algo, we minimize the dynamics error through latent state factorization, reduce the representation error via meta-state regularization, and mitigate the policy error by meta-value regularization. By integrating these three components, we effectively reduce the generalization error of \algo~on Meta-POMDPs.

\begin{table*}[t]
\centering
\begin{tabular}{
  l|cccc|cccc
}
\toprule
\multicolumn{1}{c}{} & \multicolumn{4}{c}{\textbf{(a) Train $\alpha=5, \beta=20,$ Evaluate $\alpha=5, \beta=20$}} & \multicolumn{4}{c}{\textbf{(b) Train $\alpha=5, \beta=20,$ Evaluate $\alpha=10, \beta=50$}} \\
\cmidrule(lr){2-5} \cmidrule(lr){6-9}
& {MAMBA} & {DreamerV3} & {CaDM} & {Ours} & {MAMBA} & {DreamerV3} & {CaDM} & {Ours}   \\
\midrule
Hopper  & 43.1\textpm12.7 & 52.3\textpm 12.6  & 41.4\textpm18.5  & \textbf{57.7\textpm14.2}     & 35.1\textpm18.2  & 43.3\textpm11.2  & 38.8\textpm 13.9  & \textbf{52.0\textpm 9.9}  \\
Walker & 47.1\textpm 6.8  & 56.7\textpm 12.0  & 55.2\textpm 15.8  & \textbf{60.8\textpm 7.4}    & 29.4\textpm 12.5  & 37.5\textpm 11.1 & 49.7\textpm 8.2  & \textbf{53.2\textpm 8.7} \\
Cheetah & 52.3\textpm 12.5  & \textbf{53.7\textpm 17.4} & 47.1\textpm 13.2  & 48.6\textpm 12.7  & \textbf{49.5\textpm 14.0} & 43.2\textpm 13.1  & 42.7\textpm 13.1 & 45.2\textpm 13.9 \\
\midrule

\multicolumn{1}{c}{} & \multicolumn{4}{c}{\textbf{(c) Train $\alpha=10, \beta=50,$ Evaluate $\alpha=10, \beta=50$}} & \multicolumn{4}{c}{\textbf{(d) Train $\alpha=10, \beta=50,$ Evaluate $\alpha=20, \beta=100$}} \\
\cmidrule(lr){2-5} \cmidrule(lr){6-9}
& {MAMBA} & {DreamerV3} & {CaDM} & {Ours} & {MAMBA} & {DreamerV3} & {CaDM} & {Ours}   \\
\midrule
Hopper  & 38.3\textpm 15.2 & 42.1\textpm 11.6  & 42.1\textpm13.1  & \textbf{52.1\textpm12.2}    & 21.3\textpm15.6  & 36.6\textpm25.2  & 21.2\textpm11.3  & \textbf{48.1\textpm 13.8}  \\
Walker & 29.3\textpm 18.6  & 38.2\textpm 8.9  & 29.6\textpm 14.9  & \textbf{40.1\textpm13.1}    & 16.5\textpm21.2  & 30.0\textpm 11.9 & 26.1\textpm 16.2  & \textbf{47.5\textpm 10.3} \\
Cheetah & 28.7\textpm 13.1  & 40.1\textpm 16.2 & 37.1\textpm 10.8  & \textbf{45.8\textpm9.6}     & 27.4\textpm11.6 & 41.7\textpm 21.3  & 41.2\textpm 18.6 & \textbf{44.7\textpm 11.2} \\
\toprule
\end{tabular}
\caption{Experimental results on the \textit{multi-scenarios} setting with various dynamics $T$ and reward function $r$. 
The results on the left~(a) and (c) are for the in-distribution setting, while the results on the right~(b) and (d) are for the out-of-distribution setting.
We adopt the normalized return metric with five random seeds.
}
\label{tab: question1}
\end{table*}

\begin{table*}[t]
\centering

\begin{tabular}{
  l|cccc|cccc
}
\toprule

\multicolumn{1}{c}{} & \multicolumn{4}{c}{\textbf{(a) Train $\alpha=5,$ Evaluate $\alpha=5$}} & \multicolumn{4}{c}{\textbf{(b) Train $\alpha=5,$ Evaluate $\alpha=10$}} \\
\cmidrule(lr){2-5} \cmidrule(lr){6-9}
& {MAMBA} & {DreamerV3} & {CaDM} & {Ours} & {MAMBA} & {DreamerV3} & {CaDM} & {Ours}   \\
\midrule
Hopper  & 46.2\textpm14.3 & 62.1\textpm 9.1  & 64.2\textpm15.4  & \textbf{67.2\textpm13.1}     & 31.2\textpm16.5  & 49.7\textpm12.5  & \textbf{60.2\textpm 11.2}  & 58.1\textpm 12.1  \\
Walker & 52.4\textpm 11.2  & 59.1\textpm 13.1  & 52.6\textpm 9.7  & \textbf{62.3\textpm 10.2}    & 42.0\textpm 18.8  & 35.6\textpm 13.0 & 48.2\textpm 6.9  & \textbf{59.0\textpm 11.0} \\
Cheetah & 53.0\textpm 10.8 & 48.2\textpm 11.3 & 51.2\textpm 11.3  & \textbf{56.8\textpm 9.2}  & 51.1\textpm12.3 & 45.1\textpm 16.2  & 46.3\textpm 12.6 & \textbf{49.9\textpm 15.6} \\
\midrule
\multicolumn{1}{c}{} & \multicolumn{4}{c}{\textbf{(c) Train $\alpha=10,$ Evaluate $\alpha=10$}} & \multicolumn{4}{c}{\textbf{(d) Train $\alpha=10,$ Evaluate $\alpha=20$}} \\
\cmidrule(lr){2-5} \cmidrule(lr){6-9}
& {MAMBA} & {DreamerV3} & {CaDM} & {Ours} & {MAMBA} & {DreamerV3} & {CaDM} & {Ours}   \\
\midrule
Hopper  & 51.5\textpm 12.8 & 52.6\textpm 9.1  & 61.6\textpm13.5  & \textbf{62.0\textpm11.6}     & 39.7\textpm12.6  & \textbf{49.8\textpm15.2}  & 43.8\textpm10.8  & 47.5\textpm 13.2  \\
Walker & 47.3\textpm 14.0  & 57.8\textpm 14.3  & 50.7\textpm 16.0  & \textbf{61.3\textpm8.7}    & 38.1\textpm13.7  & 44.8\textpm 10.9 & 40.7\textpm 12.7  & \textbf{49.7\textpm 15.2} \\
Cheetah & \textbf{61.2\textpm 11.6} & 52.9\textpm 16.4 & 48.3\textpm 13.8  & 60.7\textpm13.1    & 45.2\textpm14.0 & \textbf{50.1\textpm 12.9}  & 47.7\textpm 13.8 & 48.0\textpm 16.8 \\

\toprule
\end{tabular}
\caption{
Experimental results on the \textit{multi-scenarios} setting with various dynamics function $T$. 
The results on the left~(a) and (c) are for the in-distribution setting, while the results on the right~(b) and (d) are for the out-of-distribution setting.
We adopt the normalized return metric with five random seeds.
}
\label{tab: question2}
\end{table*}

\section{Experiments}
We design our experiments to answer the following questions:
(Q1) How does \algo~perform in the multi-scenario setting?
(Q2) How does \algo~perform when the dynamics function changes?
(Q3) How does \algo~perform when the observation space changes?
(Q4) What is the contribution of each proposed technique in~\algo?



\subsection{Setup}

\paragraph{Domains.}
We evaluate our method and baselines on the DMControl scenarios~\cite{tassa2018deepmind,todorov2012mujoco}.
Specifically, we select three scenarios on DMControl: hopper, walker, and cheetah. 
To change dynamics ${T}$, we modify the environment parameters by uniformly sampling the torso length and size within an interval of $\alpha\%$ around their default values. 
To change reward $r$, we adjust the agent's target speed as follows.
Let $v_{\max}$ represent the fastest speed of the agent under default dynamics settings. 
The scenario objective $v_i$ is uniformly sampled from the interval $[0, \beta\% \cdot v_{\max}]$.


\paragraph{Multi-Scenario.}
In our experiments, we train world-model capable of generalizing across diverse scenarios. 
Specifically, we merge data from Hopper, Walker and Cheetah to train the world-model. Then, we select one scenario (e.g., Hopper) for scenario adaptation. 
We conduct both in-distribution and out-of-distribution evaluations, with detailed specifications provided in the section~\ref{sec: main results}.

\paragraph{Baselines.}
To validate the effectiveness of our method, we choose the following state-of-the-art algorithms as baselines:
DreamerV3~\cite{hafner2023mastering} features a context-augmented RSSM architecture, demonstrating strong cross-domain adaptability within a unified framework.
MAMBA~\cite{rimon2024mamba} extends DreamerV3 to meta-RL through trajectory sampling strategies and an adaptive horizon scheduling mechanism.
CaDM~\cite{lee2020context} utilizes contextual trajectory data to construct bidirectional dynamics models, enabling effective generalization across diverse dynamical systems.

\begin{figure*}[t]
\centering
    \begin{subfigure}[b]{0.6\columnwidth}
      \includegraphics[width=\textwidth]{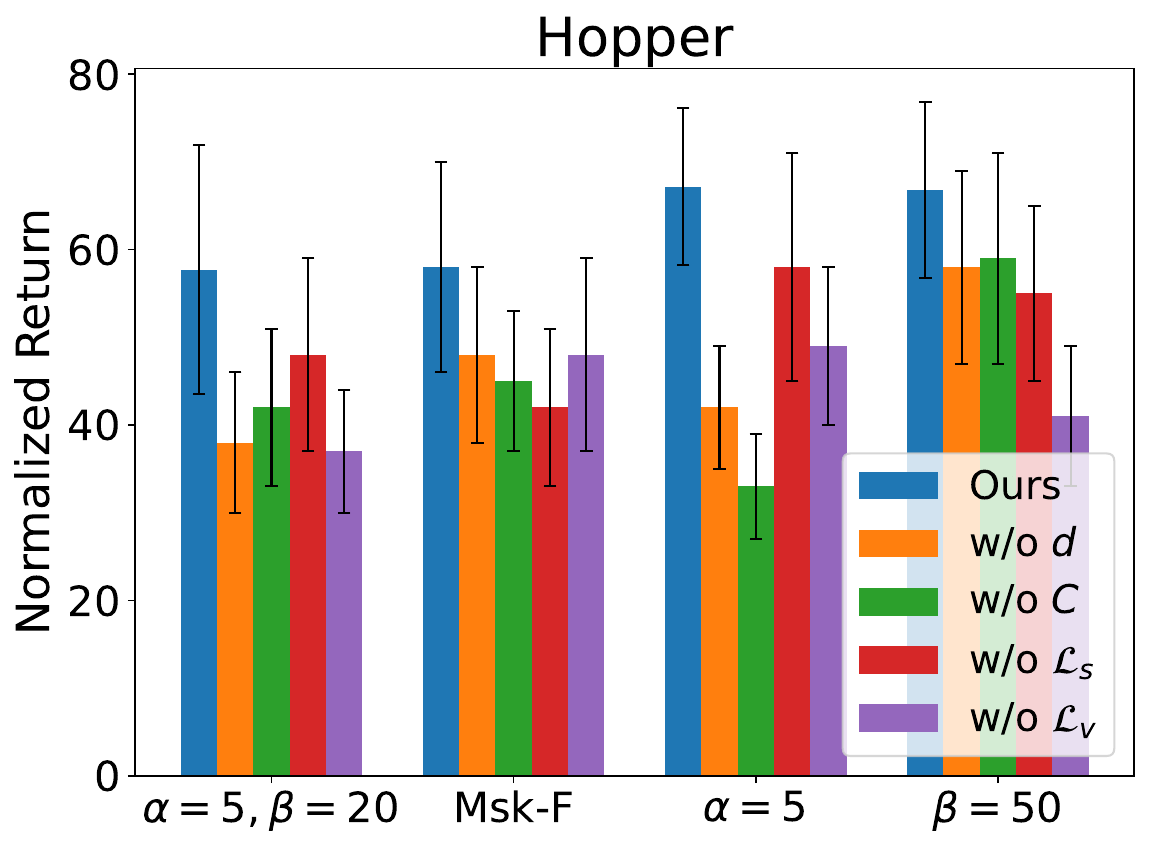}
      \label{fig:oaspl_a}
    \end{subfigure}%
    ~
    \begin{subfigure}[b]{0.6\columnwidth}
      \includegraphics[width=\textwidth]{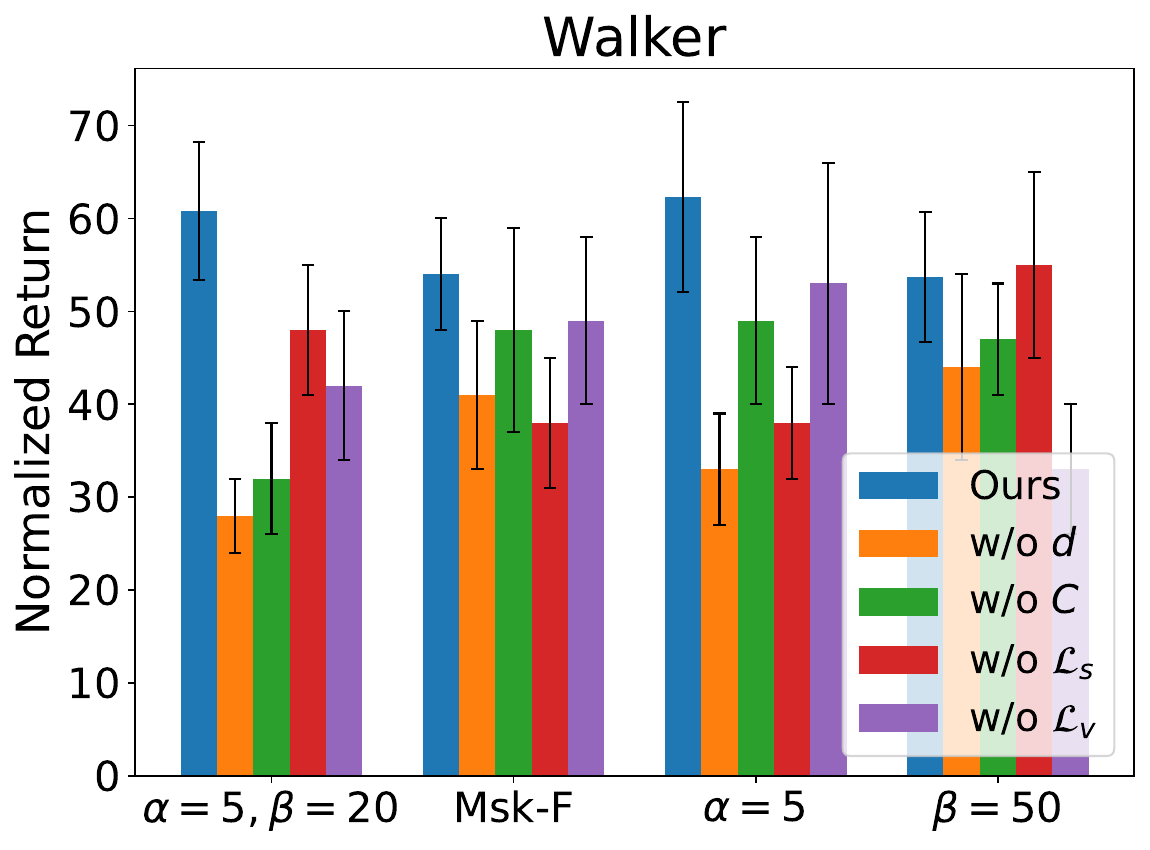}
      \label{fig:oaspl_b}
    \end{subfigure}
    \begin{subfigure}[b]{0.6\columnwidth}
      \includegraphics[width=\textwidth]{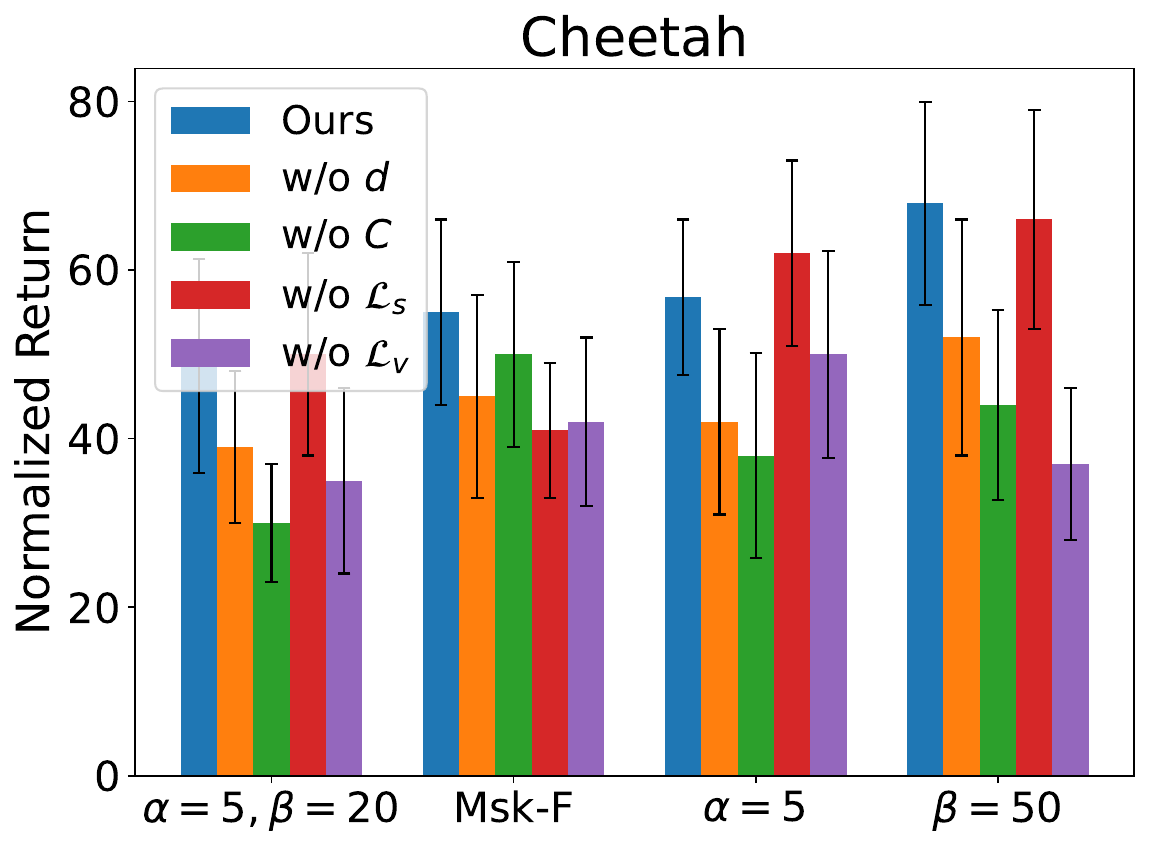}
      \label{fig:oaspl_c}
    \end{subfigure}%
\caption{Module ablation studies on various scenarios with five random seeds.}
\label{fig: ablation study}
\end{figure*}

\subsection{Main Results}
\label{sec: main results}
\paragraph{Answer for Question 1.}
In this setting, we train and evaluate algorithms in multi-scenarios by merging data from Hopper, Walker and Cheetah.
In addition, we divide the test scenarios into in-distribution and out-of-distribution.
For the in-distribution setting, the distributions of the training scenario set match those of the test scenarios~(e.g., Train $(\alpha=5,\beta=20)$, Evaluate $(\alpha=5,\beta=20)$).
For the out-of-distribution setting, the test scenario's distributions exceed those of the training set~(e.g., Train $(\alpha=5,\beta=20)$, Evaluate $(\alpha=10,\beta=50)$).


By comparing the experimental results (a) and (c) in Table~\ref{tab: question1}, we find that both~\algo~and CaDM are more adaptable to dynamic changes in meta-scenarios. \algo~demonstrates the more significant performance advantage and generalization capability over other algorithms as the range of meta-scenarios expands.
In addition, the comparison between in-distribution and out-of-distribution experimental results shows that all algorithms have a performance decline when the test distribution exceeds the training distribution.
We find MAMBA shows the most significant drop, which is attributed to the fact that the meta-scenarios do not conform to its assumption of decomposable scenario distributions. 
In contrast, our algorithm shows a smaller decline and achieves optimal results in 11 out of 12 cases.

\paragraph{Answer for Question 2.}
We fix the training scenario objective by setting the target speed to $50\%\cdot v_{\max}$.
Then, we change the dynamics function by randomly sampling parameters $\alpha$.
Like the $\textbf{Answer for Question 1}$, we train algorithms on the multi-scenarios and divide the experiments into in-distribution and out-of-distribution settings.
The experimental results in Table~\ref{tab: question2} show out-of-distribution conditions result in a performance decline for all methods, but~\algo~shows a smaller decrease, indicating stronger adaptability to out-of-distribution scenarios.
Out of 6 results, our method achieved the best performance in 5 cases, demonstrating that~\algo~provides optimal generalization performance concerning dynamics.




\paragraph{Answer for Question 3.}
In this setting, we evaluate algorithms under various observation functions $\mathcal{O}^i$ by conducting the following modifications:
$\verb|Add-G|$: Adds Gaussian noise to each dimension of the original observation.
$\verb|Add-D|$: Introduces additional noise information, equivalent to 10\% of the original dimensions, into the observation.
$\verb|Msk-R|$: Randomly masks 10\% of the observation values in the original observation.
$\verb|Msk-F|$: Masks a fixed 10\% of the observation values in the original observation across specific dimensions.
The experimental results in Table~\ref{tab: question3} show that our algorithm achieved the best results in 8 out of 12 cases across various observation functions.
Particularly in the $\verb|Add-G|$ setting, the~\algo~method demonstrates significant advantages. 

\begin{table}[t]
\begin{center}
\begin{tabular}{cccccc}
\toprule
Setting & Env & MAMBA &  DV3 & CaDM & Ours  \\
\hline
\multirow{4}{*}{Add-G}  & Hopper    & 63.7 & 59.1 & 49.3 & \textbf{65.3}  \\
& Walker      & 59.1 & 51.2 & 51.7 & \textbf{63.1} \\ 
& Cheetah     & 49.6 & 42.7 & 48.2 & \textbf{59.3}  \\
\hline
\multirow{4}{*}{Add-D}   & Hopper   & 68.2 & \textbf{69.7} & 58.2 & 61.3  \\
& Walker      & 60.1 & 55.2 & 54.8 & \textbf{60.7} \\ 
& Cheetah     & 50.7 & \textbf{59.2} & 44.9 & 58.6  \\
\hline
\multirow{4}{*}{Msk-R}   & Hopper   & 71.2 & \textbf{72.3} & 64.3 & 70.1  \\
& Walker      & 63.3 & 59.6 & 61.2 & \textbf{64.2} \\ 
& Cheetah     & 58.2 & 60.2 & 49.3 & \textbf{63.5}  \\
\hline
\multirow{4}{*}{Msk-F}   & Hopper   & 50.1 & 48.2 & 39.6 & \textbf{58.6}  \\
& Walker      & 48.7 & 51.3 & 57.1 & \textbf{54.1} \\ 
& Cheetah     & \textbf{61.0} & 57.1 & 49.3 & 55.6  \\
\bottomrule
\end{tabular}
{\caption{
Experimental results with various observation functions.
We adopt the normalized return metric with five random seeds.
}
\vspace{-10pt}
\label{tab: question3}}
\end{center}
\end{table}

\subsection{Ablation Study}
This experiment primarily examines the impact of various modules of our method on the final performance and validates their implementation's effectiveness on the overall model. Our experiments are divided into the following groups:
(1) w/o $d$: Without constructing the deterministic state $d$, using only the stochastic states $u, h$.
(2) w/o $C$: Without using context and prompt scenario information during state prediction.
(3) w/o $\mathcal{L}_s$: Without the meta-state regularization loss.
(4) w/o $\mathcal{L}_v$: Without constructing the meta-value and using meta-value regularization during learning the dynamics model.

The experimental settings also incorporate the aforementioned different configurations to compare each component's impact comprehensively. 
The experimental results in Figure~\ref{fig: ablation study} show that w/o $\mathcal{C}$ has the most significant impact in the multi-scenario setting. 
The absence of $d$ considerably affects temporal information, thus significantly impacting dynamics changes.
Overall, it can be seen that the complete~\algo~consistently achieves optimal performance across different settings.

\section{Related works}

\paragraph{Model-based RL} offers an effective methods to enhance sampling efficiency~\cite{nagabandi2018learning}. Dyna~\cite{sutton1991dyna} uses data generated from model-based predictions for data augmentation in agent trajectories. 
Muzero~\cite{schrittwieser2020mastering}, on the other hand, employs Monte Carlo Tree Search ~\cite{coulom2006efficient} based on its model, enabling the identification of superior strategies with fewer interactions with the environment. 
However, previous model-based methods often focus on generalization concerning changes in dynamics~\cite{lee2020context} or adapting to diverse reward functions~\cite{rimon2024mamba,kidambi2020morel}, without addressing the issue of unified modeling across different scenarios. Additionally, there is an increasing focus on resolving the challenge of learning models that align to acquire an effective policy~\cite{lambert2020objective}. 
This can be primarily categorized into two approaches: value-aware~\cite{abachi2020policy,wang2023live} and policy-aware~\cite{grimm2021proper,voelcker2022value,voloshin2021minimax}. However, these methods often make world models more task-specific, thereby losing the ability to generalize across scenarios. 

\paragraph{Meta RL} enables agents to learn from related tasks and then adapt to new tasks with minimal interaction data. Most meta-RL algorithms adhere to a model-free paradigm, subdivided into optimization-based and context-based methodologies. MAML~\cite{finn2017model} and PrOMP~\cite{rothfuss2018promp}, exemplifying optimization-based meta-RL, update policies based on gradients derived from training tasks. Context-based meta-RL, represented by RL2~\cite{duan2016rl}, suggests conditioning policies on hidden states extracted from task trajectories. PEARL~\cite{rakelly2019efficient} distinguishes task inference from action selection, facilitating explicit task representations. 
Unlike that, in this paper, we hope to learn a world model that can be generalized across multiple scenarios.

\section{Conclusion}

In this paper, we aim to improve the generalization ability of the world-model across multi-scenarios.
To address this issue, we propose~\algo, a unified world-model capable of generalizing across different scenarios.
Specifically, \algo~adopts meta-state regularization to extract unified representation of scenario-relevant information, and meta-value regularization to align world-model optimization with policy learning across diverse scenario objectives.
We theoretically analyze the upper bound of the world-model's generalization error.
We conduct extensive experiments using the MuJoCo-based physics engine.
The experimental results show that the learned world-model achieves strong generalization ability across multi-scenarios.

\bibliography{aaai2026}

\clearpage
\onecolumn
\appendix
\clearpage

\section{Algorithm}
\label{app: alg}
We present the algorithm pseudocode of MrCoM, which is divided into two phases: world-model training and scenario adaptation, as shown in Algorithm~\ref{alg:train} and Algorithm~\ref{alg:test}, respectively.

\begin{algorithm}[H]
\caption{World-Model Training}
\label{alg:train}
\textbf{Input}: Meta tasks set $p(\mathcal{T})$ \\
\textbf{Output}: World-model $\hat{T}_\theta$  
\begin{algorithmic}[1] 
\STATE Initialize world-model parameters $\theta$, meta-value parameters $\psi$ and task-value parameters $\{\psi_i\}$
\WHILE{not converged}
    \FOR{$N$ step}
        \STATE Sample task $\mathcal{T}^i \sim p(\mathcal{T})$;
        \STATE Sample trajectories $\{(o, a, o^\prime)\}$ by $\pi_{\phi_i}$ through $\mathcal{T}^i$;
        \STATE Update $\psi_i \leftarrow  \psi_i - \alpha \frac{d}{d \psi_i} \mathcal{L}_{\text{value}_i}$ based on Equation~\ref{equ:loss1};
        \STATE Update $\phi_i$ based on standard RL algorithms;
        \STATE Sample datapoints  $ (o, v_{\psi_i}(\tilde{s}))$ and add to  $\mathcal{D}(\mathcal{T}^i)$;
    \ENDFOR
    \STATE Update $\psi \leftarrow  \psi - \lambda \frac{d}{d \psi} \sum_{\mathcal{T}^i \sim p(\mathcal{T})} \mathcal{L}_\text{value}$ based on Equation~\ref{equ:loss2};
    \WHILE{not converged}
        \STATE Calculate $\mathcal{L}_\text{MrCoM}$ in Equation~\ref{loss:mrcom};
        \STATE Update $\theta \leftarrow \theta -\beta \frac{d}{d\theta} \mathcal{L}_\text{MrCoM}$;
    \ENDWHILE
\ENDWHILE
\end{algorithmic}
\end{algorithm}

\begin{algorithm}[H]
\caption{Scenario Adaptation}
\label{alg:test}
\textbf{Input}: World-model $\hat{T}_\theta$, dataset $\mathcal{D}$, target task $\mathcal{T}^{\text{target}}$ \\
\textbf{Output}: Policy $\pi_\phi$ 
\begin{algorithmic}[1] 
\STATE Initialize policy $\pi_\phi$ and model datasets $\mathcal{D}_{\text{model}}$
\WHILE{not converged}
    \FOR{$E$ steps}
        \STATE Take action based on $\pi_\phi$ in $\mathcal{T}^{\text{target}}$ and add to $\mathcal{D}$;
        \FOR{$H$ rollouts horizon}
            \STATE Sample $o$ from $\mathcal{D}$;
            \STATE Sample $o^\prime, r$ through world-model $\hat{T}_{\theta}$;
            \STATE Add $(o, a, r, o^\prime)$ to $\mathcal{D}_{\text{model}}$;
        \ENDFOR
        \FOR{$G$ policy updates}
            \STATE Sample $(o, a, r, o^\prime)$ from $\mathcal{D} \cup \mathcal{D}_{\text{model}}$;
            \STATE Update $\psi$ and $\phi$ based on  standard RL algorithms;
        \ENDFOR
    \ENDFOR
    \FOR{not converged}
        \STATE Sample trajectories $\{(o, a, o^\prime)\}$ from $\mathcal{D}$;
        \STATE Calculate $\mathcal{L}_\text{MrCoM}$ in Equation~\ref{loss:mrcom};
        \STATE Update $\theta \leftarrow \theta -\beta \frac{d}{d\theta} \mathcal{L}_\text{MrCoM}$;
    \ENDFOR
\ENDWHILE
\end{algorithmic}
\end{algorithm}
\clearpage
\section{Proof}
\label{appendix: proof}

\subsection{Proof of Lemma~\ref{state_dyna}}
\label{proof: lemma3}
\begin{proof}
\begin{align}
&D_{TV}  \left( \widetilde{T} \left(f(o^\prime) \mid f(o), a \right) \| T^{i}\left(s^{\prime} \mid s, a \right)\right)~\nonumber  \\
&= D_{TV}\left(\widetilde{T}\left(s^\prime + \delta s^\prime \mid s+\delta s , a\right) \| T^{i}\left(s^{\prime} \mid s, a \right)\right)   ~\label{eq_state_dyna1}  \\ 
&\leq D_{T V}\left(\widetilde{T}\left(s^\prime \mid s, a\right) \| T^{i}\left(s^{\prime} \mid s, a\right)\right) ~\nonumber + \frac{1}{2} \sum\left(\delta s^\prime \cdot \nabla_{s^\prime} T+\delta s \cdot \nabla_{s} T\right)  ~\label{eq_state_dyna2} \\
&\leq \epsilon_{T} + \frac{1}{2} \sum\left(\epsilon_S \cdot \nabla_{s^\prime} T+\epsilon_S \cdot \nabla_{s} T\right) \\
&\leq \epsilon_T + C_{T} \cdot \epsilon_{s} \label{eq_state_dyna3} 
\end{align}

Equations \ref{eq_state_dyna1} to \ref{eq_state_dyna2} employ a Taylor expansion. In Equation \ref{eq_state_dyna3}, $C_T$ denotes the maximum derivative of the dynamics function with respect to $s$, representing the sensitivity of state changes to the state transition function. Specifically, $C_T = \max _s \nabla_s \sum_a T(s^\prime \mid s, a)$. Furthermore, the second-order terms can be neglected, leading to the approximate derivation.
\end{proof}

\subsection{Proof of Lemma~\ref{state_pi}}
\label{proof: lemma4}

\begin{proof}
\begin{align}
&D_{TV}\left(\pi_{1}(a \mid f(o)) \| \pi_{2}(a \mid s)\right) \nonumber \\
&= D_{TV}\left(\pi_{1}(a \mid s + \delta s) \| \pi_{2}(a \mid s )\right) \\
&\leq D_{T V}\left(\pi_{1}(a \mid s) \| \pi_{2}(a \mid s)\right) \label{eq_policy_dyna1} + \frac{1}{2} \epsilon_S \sum \nabla_{s} \pi_1(a|s) \\
&\leq \epsilon_{\pi} + \frac{1}{2}C_{\pi} \label{eq_policy_dyna2} \cdot  \epsilon_S 
\end{align}
In Equation \ref{eq_policy_dyna1}, we perform a Taylor expansion on the policy difference, considering the error in the input state and neglecting higher-order terms. In Equation \ref{eq_policy_dyna2}, we introduce the upper bound of state representation error $\epsilon_S$ and quantify it using the sensitivity of the policy function to state variations, $C_{\pi} = \max _s \nabla_s \sum _a \pi_1(a|s)$.
\end{proof}

\subsection{Proof of Lemma~\ref{dyna_pi}}
\label{proof: lemma5}
The proof of Lemma~\ref{dyna_pi} follows from the derivation presented in~\cite{janner2019trust}. Subsequently, we present Lemma~\ref{TVD} and Lemma~\ref{MTVD} without intermediate derivations.

\begin{lemma}~\label{TVD}
TV Distance of Joint Distributions: Suppose we have two distributions $p_1(x, y) = p_1(x)p_1(y|x)$ and $p_2(x, y) = p_2(x)p_2(y|x)$. We can bound the total variation distance of the joint as:
\begin{align}
D_{T V}\left(p_{1}(x, y) \| p_{2}(x, y)\right) \leq D_{T V}\left(p_{1}(x) \| p_{2}(x)\right)
+\max _{x} D_{T V} \left(p_{1}(y \mid x) \| p_{2}(y \mid x)\right)
\end{align}
\end{lemma}

\begin{lemma}~\label{MTVD}
Suppose the expected KL-divergence between two transition distributions is bounded as $\max _{t} D_{K L}\left(p_t^{1}\left(s^{\prime} \mid s\right) \| p_t^{2}\left(s^{\prime} \mid s\right)\right) \leq \delta$, and the initial state distributions are the same $\rho_0^1=\rho_0^2$. Then the distance in the state marginal is bounded as:
\begin{align}
    D_{T V}\left(p^{1}_{t}(s) \| p^{2}_{t}(s)\right) \leq t \delta
\end{align}
\end{lemma}


\begin{proof}
Let $G^1(\pi_1)$ denote returns of $\pi_1$ under dynamics $T^1$, and $G^2(\pi_2)$ denote returns of $\pi_2$ under dynamics $T^2$. 
By substituting the relationship between the state transition probability, transition function, and policy function, $p_t(s^{\prime} \mid s) = p_t(s^{\prime} \mid s, a) \cdot p_t(a \mid s)$, into Lemma~\ref{TVD}, we obtain
\begin{equation}
    D_{KL}\left(p_t^{1}\left(s^{\prime} \mid s\right) \| p_t^{2}\left(s^{\prime} \mid s\right)\right) \leq \epsilon_T+\epsilon_\pi
\end{equation}

We can then proceed with the following derivation:
\begin{align}
& D_{T V}\left(p^{1}_{t}(s,a) \| p^{2}_{t}(s,a)\right) \\
& \leq D_{T V}\left(p^{1}_{t}(s) \| p^{2}_{t}(s)\right) + D_{T V}  \left(p^{1}_{t}(a \mid s) \| p^{2}_{t}(a \mid s)\right)  \label{tvd1} \\
& \leq D_{T V}\left(p^{1}_{t}(s) \| p^{2}_{t}(s)\right) + \epsilon_\pi \label{tvd2} \\
& \leq t(\epsilon_T+\epsilon_\pi) + \epsilon_\pi \label{tvd3}
\end{align}
Equation~\ref{tvd1} utilises Lemma~\ref{TVD}, and Equation~\ref{tvd3} employs Lemma~\ref{MTVD}. Next, we proceed with the derivation of the policy performance.
\begin{align}
& \left|G^1(\pi_1)-G^2(\pi_2)\right| =\left|\sum_{s, a}\left(p^{1}(s, a)-p^{2}(s, a)\right) r(s, a)\right| \\
& =\left|\sum_{s, a}\left(\sum_{t} \gamma p^{1}_{t}(s, a)-p^{2}_{t}(s, a)\right) r(s, a)\right| \\
& =\left|\sum_{t} \sum_{s, a} \gamma \left(p^{1}_{t}(s, a)-p^{2}_{t}(s, a)\right) r(s, a)\right| \\
& \leq \sum_{t} \sum_{s, a} \gamma \left|p^{1}_{t}(s, a)-p^{2}_{t}(s, a)\right| r(s, a) \\
& \leq R \sum_{t} \sum_{s, a} \gamma \left|p^{1}_{t}(s, a)-p^{2}_{t}(s, a)\right|
\end{align}

Substituting Equation~\ref{tvd3} into the above expression yields:
\begin{align}
&\left|G^1(\pi_1)-G^2(\pi_2)\right| \\
&\leq 2R \sum_t \gamma D_{T V}\left(p^{1}_{t}(s,a) \| p^{2}_{t}(s,a)\right) \\
&\leq 2 R \sum_{t} \gamma t\left(\epsilon_{m}+\epsilon_{\pi}\right)+\epsilon_{\pi} \\
&\leq \frac{2 R \gamma\left(\epsilon_{\pi}+\epsilon_{T}\right)}{(1-\gamma)^{2}}+\frac{2 R \epsilon_{\pi}}{1-\gamma}
\end{align}
\end{proof}

\subsection{Proof of Theorem~\ref{theorem2}}
\label{proof theorem 2}
\begin{proof}
We first derive the upper bound of the performance difference under state representation error.
We denote $\widetilde{G}^1(\pi_1)$ as the performance of $\pi_1(f(o))$ under the dynamics $\widetilde{T}^1(f(o), a)$.
Let $G^2(\pi_2)$ represent the performance of $\pi_2(s)$ under the dynamics $T^2(s, a)$. 
Given dynamics model error $\epsilon_T$, policy difference in the state space $\epsilon_\pi$, and the state representation error $\epsilon_{S}$, the bound of the policy performance can be determined as: 
\begin{align}
        \left|\widetilde{G}^1(\pi_1) - G^2(\pi_2)\right| \leq \frac{ R\gamma[2\epsilon_{\pi}+ 2\epsilon_{T} +(C_\pi +2C_T) \epsilon_{S} ] }{(1-\gamma)^{2}}  
        +\frac{R (2\epsilon_{\pi}+ C_\pi \epsilon_{S})}{1-\gamma}.
\end{align}

Ultimately, we aim to derive the upper bound on the difference between the performance of the same policy in the learned dynamics environment and the true environment. 
We can bridge this gap by following \cite{janner2019trust} and introducing a behavior policy $\pi_D$ to relate $\widetilde{G}_\theta(\pi)$ and $\widetilde{G}^i(\pi)$:

\begin{equation}
    \left|\widetilde{G}^i(\pi)-\widetilde{G}_\theta(\pi)\right|=\underbrace{\widetilde{G}^i(\pi)-G^i\left(\pi_{D}\right)}_{L_{1}}+\underbrace{G^i\left(\pi_{D}\right)-\widehat{G}_\theta(\pi)}_{L_{2}},
\end{equation}
where $\widetilde{G}_\theta(\pi)$ denotes the performance under the learned model $\hat{T}_\theta$, while $\widetilde{G}^i(\pi)$ represents the performance in the true environment of task $\mathcal{T}^i$. 
$G^i\left(\pi_{D}\right)$ represents the performance of the behavioral policy in the true environment without representation error.
For $L_1$, since both terms are in the real environment with no model errors, we have the following conclusion:
\begin{equation}
L_1 \leq \frac{ R\gamma(2\epsilon_{\pi} ) }{(1-\gamma)^{2}} 
        +\frac{R (2\epsilon_{\pi}+ C_\pi \epsilon_{S})}{1-\gamma}.
\end{equation}
For $L_2$, we have:
\begin{equation}
L_2 \leq  \frac{ R\gamma[2\epsilon_{\pi}+ 2\epsilon_{T} +(C_\pi +2C_T) \epsilon_{S} ] }{(1-\gamma)^{2}}  \\
        +\frac{R (2\epsilon_{\pi}+ C_\pi \epsilon_{S})}{1-\gamma}.
\end{equation}
By adding $L_1$ and $L_2$, we can obtain the generalization bound in the Meta-POMDP setting:
\end{proof}
\clearpage
\section{Additional Experiments}
\label{appendix: addition exp}

\subsection{Experiments in the Single Scenario Setting}
We conduct experiments in the single scenario setting with changing hyperparameters $\alpha$ and $\beta$.
The experimental results in Table~\ref{exp1-2} show that while our algorithm's advantage is less pronounced than the multi-scenario setting, it achieves optimal performance in 6 out of 12 results, demonstrating its effectiveness, especially in OOD scenarios.




\begin{table}[H]
\centering
\begin{tabular}{
  l|cccc|cccc
}
\toprule
\multicolumn{1}{c}{} & \multicolumn{4}{c}{\textbf{(a) Train $\alpha=5, \beta=20$, Evaluate $\alpha=5, \beta=20$}} & \multicolumn{4}{c}{\textbf{(b) Train $\alpha=5, \beta=20$, Evaluate $\alpha=10, \beta=50$}} \\
\cmidrule(lr){2-5} \cmidrule(lr){6-9}
& {MAMBA} & {DreamerV3} & {CaDM} & {Ours} & {MAMBA} & {DreamerV3} & {CaDM} & {Ours}   \\
\midrule
Hopper  & 47.2\textpm11.3 & 54.7\textpm 11.2  & 48.3\textpm12.1  & \textbf{58.2\textpm15.3}     & 42.7\textpm12.1  & \textbf{53.9\textpm12.0}  & 43.2\textpm 14.1  & 53.1\textpm 13.5  \\
Walker & 58.2\textpm 10.8  & \textbf{65.3\textpm 14.2}  & 63.1\textpm 14.2  & 62.3\textpm 13.7    & 36.8\textpm 18.5  & 48.2\textpm 12.1 & 48.0\textpm 10.1  & \textbf{54.1\textpm 14.2} \\
Cheetah & 61.7\textpm 12.8  & \textbf{62.3\textpm 9.1} & 57.2\textpm 12.5  & 56.6\textpm 11.2  & 50.1\textpm 13.6 & 55.4\textpm 15.2  & 52.1\textpm 16.4 & \textbf{56.4\textpm 11.9} \\
\midrule

\multicolumn{1}{c}{} & \multicolumn{4}{c}{\textbf{(c) Train $\alpha=10, \beta=50$, Evaluate $\alpha=10, \beta=50$}} & \multicolumn{4}{c}{\textbf{(d) Train $\alpha=10, \beta=50$, Evaluate $\alpha=20, \beta=100$}} \\
\cmidrule(lr){2-5} \cmidrule(lr){6-9}
& {MAMBA} & {DreamerV3} & {CaDM} & {Ours} & {MAMBA} & {DreamerV3} & {CaDM} & {Ours}   \\
\midrule
Hopper  & 42.1\textpm 13.3 & 48.4\textpm 10.2  & \textbf{54.2\textpm14.7}  & 52.7\textpm11.8    & 28.4\textpm13.6  & 42.1\textpm19.7  & 31.7\textpm13.9  & \textbf{45.3\textpm 15.6}  \\
Walker & 32.0\textpm 16.4  & 43.2\textpm 15.9  & 33.7\textpm 13.8  & \textbf{44.9\textpm12.1}    & 18.9\textpm16.2  & 37.6\textpm 12.9 & 32.8\textpm 13.1  & \textbf{45.2\textpm 13.3} \\
Cheetah & 32.1\textpm 14.2  & \textbf{47.6\textpm 17.2} & 36.8\textpm 14.9  & 46.2\textpm13.1     & 33.8\textpm16.0 & \textbf{48.3\textpm 18.2}  & 46.4\textpm 13.2 & 43.1\textpm 17.2 \\
\toprule
\end{tabular}
\caption{Experimental results on the single-scenario setting with various dynamics. 
The results on the left~(a) and (c) are for the in-distribution setting, while the results on the right~(b) and (d) are for the out-of-distribution setting.
We adopt the normalized return metric with five random seeds.}~\label{exp1-2}
\end{table}

\subsection{Tasks with Different Behavior Policies}
In this experiment, we aim to evaluate how the behavior policy used for data sampling affects the model's adaptability. 
Specifically, we set the environment hyperparameters $\alpha=0$ and $\beta=50$, meaning the task objectives are randomly sampled while the dynamics function remains fixed.
Then, we first collect offline data using different behavior policies $\pi_D$, pre-train the dynamics model, and then adapt it to a specific task. Adaptation is fixed at 100k interaction steps with the environment. 
The data comes from behavior policies with environment hyperparameter $\beta=0$, targeting $50\cdot \%v$. The four datasets are described as follows:
\begin{enumerate}
    \item Expert: Data from the highest return behavior policy.
    \item Medium: Data from a behavior policy with half the maximum return.
    \item Mix: Data from behavior policies during the training process.
    \item Random: Data from a random behavior policy.
\end{enumerate}

The experimental results in Table~\ref{exp_policy} show that our algorithm achieves optimal performance in 7 out of 12 results. 
It demonstrates less dependency on behavior policies. While Dreamer and MAMBA perform well with expert data due to their design for online learning, CaDM, which addresses changes in dynamics functions, shows no advantage when the objective changes. This experiment confirms that \algo~has stronger adaptability to different behavior policies.

\begin{table}[H]
\begin{center}
\begin{tabular}{cccccc}
\toprule
Setting & Env & MAMBA &  Dreamer & CaDM & Ours  \\
\hline
\multirow{3}{*}{Expert}   & Hopper      & 59.3 & 63.2 & 49.7 & \textbf{66.8}  \\
                      & Walker          & 53.2 & \textbf{59.4} & 41.0 & 53.7 \\ 
                      & Cheetah         & \textbf{69.2} & 64.7 & 58.2 & 67.9   \\
\hline
\multirow{3}{*}{Medium}   & Hopper      & 48.2 & 53.1 & 44.5 & \textbf{59.8} \\
                      & Walker          & 52.3 & 58.5 & 46.1 & \textbf{61.2} \\ 
                      & Cheetah         & 44.7 & 51.8 & 39.7 & \textbf{52.6}   \\
\hline
\multirow{3}{*}{Mix}   & Hopper         & 41.1 & \textbf{54.7} & 39.6 & 50.1 \\
                      & Walker          & 38.2 & 49.0 & 36.8 & \textbf{52.9} \\ 
                      & Cheetah         & 33.4 & 46.8 & 30.1 & \textbf{47.4} \\
\hline
\multirow{3}{*}{Random}    & Hopper     & 38.3 & 42.1 & 42.7 & \textbf{43.2} \\
                      & Walker          & 27.3 & \textbf{41.6} & 35.2 & 39.1 \\ 
                      & Cheetah         & 34.6 & 36.8 & \textbf{41.8} & 37.2 \\
\bottomrule
\end{tabular}
{\caption{Results of tasks with different behavior policies}\label{exp_policy}}
\end{center}
\end{table}

\clearpage
\section{Experimental Details}
\label{appendix: details}
The detailed hyperparameters of MrCoM are shown in table \ref{tab:htsp}.
We will release our code after the paper is accepted to help readers reproduce our experiments.

\begin{table}[h]
  \centering
\begin{tabular}{lc}
\toprule
\textbf{Name}     & \textbf{Value} \\
\midrule
world-model \\
\midrule
    Loss scale $\lambda_s$           &  0.1  \\
    Loss scale $\lambda_\text{var}$  &  1  \\
    Loss scale $\lambda_v$           &  1  \\
    Batch size $B$        &  32 \\
    Learning rate    & 1e-4 \\
    Learning rate decay weight & 1e-4\\
    MLP layers of tokenizer & 3 \\
    Activation  & ReLU \\
    Embedding dimension  & 128 \\
    Prompt length    &    20\\
    Layers of transformer & 1 \\
    Deterministic state dimension & 128 \\
    Stochastic state dimension & 128 \\
    Number of attention heads & 3\\
    MLP layers of reward module & 3\\
    MLP layers of observation decoder & 3 \\
\midrule
policy \\
\midrule
Discount $\gamma$ & 0.99\\
Actor learning rate & 1e-4 \\
Critic learning rate & 2e-4 \\
Horizon $H$ & 5\\
Policy updates $G$ & 10 \\

\bottomrule
\end{tabular}%
\caption{Hyperparameters of MrCoM}
\label{tab:htsp}
\end{table}

\twocolumn
\clearpage
\input{checklist}
\end{document}